\documentclass{article}

\PassOptionsToPackage{numbers, compress}{natbib}

\usepackage{amsmath}
\usepackage{algpseudocode}
\usepackage{algorithm}
\usepackage{url}
\usepackage[utf8]{inputenc}
\usepackage[T1]{fontenc}
\usepackage{multirow}
\usepackage{booktabs}
\usepackage{makecell}

\usepackage[preprint]{neurips_2025}

\usepackage[utf8]{inputenc} %
\usepackage[T1]{fontenc}    %
\usepackage{hyperref}       %
\usepackage{url}            %
\usepackage{booktabs}       %
\usepackage{amsfonts}       %
\usepackage{nicefrac}       %
\usepackage{microtype}      %
\usepackage{xcolor}         %
\usepackage{xspace}

\usepackage{amsmath}
\usepackage{amssymb}
\usepackage{mathtools}
\usepackage{amsthm}
\usepackage{bm}
\usepackage{arydshln}
\usepackage[most]{tcolorbox}
\usepackage{array} 
\usepackage{mdframed}
\usepackage{siunitx}
\usepackage{wrapfig}

\usepackage{caption}
\usepackage[most]{tcolorbox}

\tcbset{
  finding style/.style={
    enhanced,
    colback=gray!10,           %
    colframe=gray!60,         %
    boxrule=0.8pt,             %
    arc=2mm,                   %
    left=1mm, right=1mm,       %
    top=1mm, bottom=1mm,
    before skip=10pt, after skip=10pt,
    fontupper=\normalfont,     %
    fonttitle=\normalfont\bfseries, %
  }
}

\newtcolorbox{findingbox}[1][]{finding style,#1}

\usepackage{algorithm}
\usepackage{algpseudocode}

\usepackage{amsmath}

\theoremstyle{plain}

\theoremstyle{definition}

\theoremstyle{remark}

\usepackage{enumitem}

\usepackage{bm}
\newcommand\ours{\texttt{CoLa}\xspace}

\title{Skip a Layer or Loop it?\\
Test-Time Depth Adaptation of Pretrained LLMs}

\author{
Ziyue Li$^{1}$, Yang Li, Tianyi Zhou\\
$^{1}$Department of Computer Science, University of Maryland, College Park \\
\texttt{litzy619@umd.edu, yli.ml.research@gmail.com, tianyi.david.zhou@gmail.com}
}

\begin{document}

\maketitle

\begin{abstract}
    \textit{Can a pretrained neural network adapt its architecture to different inputs without any finetuning? Do we need all layers for simple tasks, and are they adequate for challenging tasks?}
    We found that the layers of a pretrained large language model (LLM) can be manipulated as separate modules to build a better and even shallower model customized for each test sample. 
    In particular, each layer from the pretrained model can be skipped/pruned or repeated multiple times as recurrent neural networks (RNN), and stacked with others in arbitrary orders, yielding a chain-of-layers (\ours) per sample. 
    This compositional space greatly expands the scope of existing works on looped/recurrent pretrained modules, layer pruning, or early-exit networks. 
    We develop a Monte Carlo Tree Search (MCTS) protocol to explore and identify the optimal \ours for each sample from math and commonsense reasoning benchmarks. 
    Compared to a static model of a fixed depth, \ours allows shortcut paths (fast thinking), recurrence of the same layer(s) (slow thinking), and combining both, offering more flexible, dynamic architectures for different inputs. 
    We conduct an extensive analysis of the MCTS-optimized \ours, which leads to two key findings:
    (1) For $>$75\% of samples with \textit{correct predictions} by the original LLM, we can find shorter \ours, suggesting a large space for improving inference efficiency; 
    (2) For $>$60\% of samples with \textit{originally incorrect predictions}, we can identify \ours achieving correct predictions, suggesting a large space of performance enhancement. 
    Our results highlight the shortcomings of using a fixed architecture of pre-trained LLMs for inference on different samples and pave the way to unlock the generalization power of test-time depth adaptation. \looseness-1
\end{abstract}

\section{Introduction}

Modern Transformer architectures, such as large language models (LLMs), have demonstrated unprecedented generalization capabilities on diverse downstream tasks such as reading comprehension and reasoning~\cite{rae2021scaling}. However, their architectures always stay the same during inference for all different tasks and test samples, despite their difference in difficulty, complexity, and distribution gap from the training task/data~\cite{peng2024limitations,dziri2023faith}. Is it necessary to apply all layers for simple tasks? On the contrary, is the pretrained model deep enough to address challenging tasks that require sophisticated reasoning? These raise the question regarding a new dimension of model generalization: \textbf{without any further training}, can a pretrained neural network's layers adapt to each task or sample by composing a model with a different depth? Answers to this question are also critical to investigating the human-like fast-slow thinking capability of pretrained LLMs: \textit{By skipping, repeating, and rearranging their layers, can we dynamically adjust the depth and architecture of the model for each task?}

Existing works on test-time rearrangement of layers mainly focus on a limited scope of depth adaptation, e.g., early-exit neural networks~\cite{teerapittayanon2016branchynet,cambazoglu2010early,liu2021anytime} or layer pruning~\cite{fan2019reducing} that aims to remove the redundant layers. They are inspired and supported by the similarity analysis of representations across layers, a potential effect of residual connections in Transformers. On the other hand, to extend the Transformer architecture to recurrent neural networks (RNN)~\cite{dehghani2018universal}, recent works such as looped transformer~\cite{fan2024looped} and recurrent depth~\cite{geiping2025scaling} explore the potential of each layer or the whole model as an RNN cell. By repeating the same pretrained layer(s) during inference, the architecture may perform slower and deeper thinking than on training data, or handle inputs of unseen lengths. In addition, it has not been studied to change the order of selected layers for test-time adaptation, which can provide a more flexible space for the architecture search. 

\begin{figure}[h]
    \begin{center}
    \vspace{-10pt}
    \includegraphics[width=0.8\textwidth]{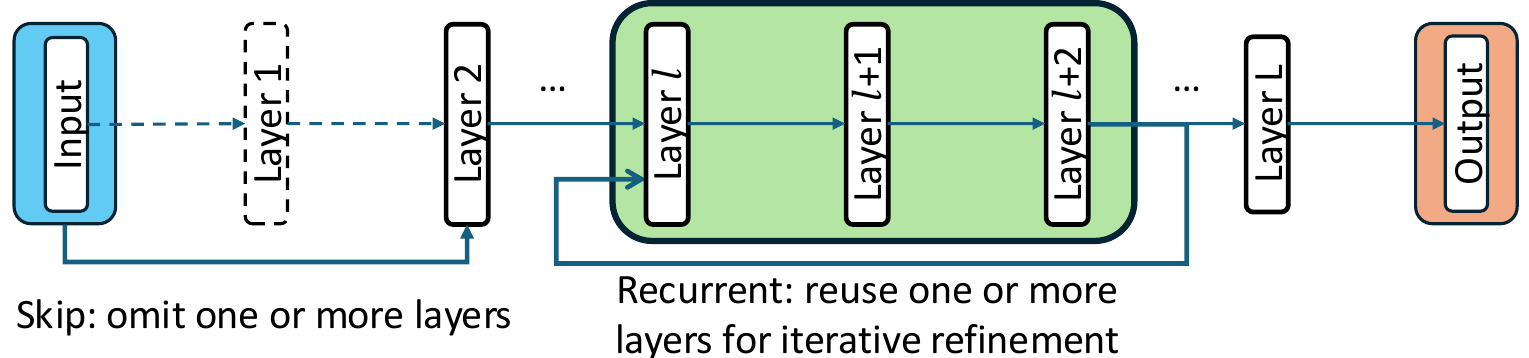}
  \end{center}
      \caption{\textbf{Test-time layer composition search space} for \ours. Starting from the original forward path, each input can dynamically skip or recurrently reuse any layer(s) to construct a custom Chain-of-Layers (\ours). This joint space enables both layer pruning and recurrence, supporting fast-slow depth adaptation and dynamic architecture generalization from a pretrained model without any finetuning.\looseness-1
}
\vspace{-8pt}
  \label{fig:search_space}
\end{figure}

In this paper, we extend the scope of depth adaptation and architecture generalization to a search space that allows selecting, skipping, and repeating arbitrary layers when constructing a customized chain-of-layers (\ours) model out of a pretrained LLM, as shown in Figure~\ref{fig:search_space}. The space enables us to remove redundant layers for each sample and integrate Transformer and RNNs architectures in one model. Hence, it offers more flexibility to process different samples with a varying number of layers. Specifically, we conduct an extensive empirical study to verify that there often exists a better (smaller or more accurate, or both) \ours than the original one for many test samples/tasks. To this end, we propose a principal and efficient Monte-Carlo Tree search (MCTS) protocol to search \ours with better prediction and/or shallower depth (i.e., minimum necessary) than the original model for each sample. Specifically, MCTS starts from an initial \ours and edits it by skipping or repeating layers for multiple rounds to find \ours that maximizes an upper confidence bound (UCB) objective, which performs an exploitation-exploration trade-off with a depth penalty. 

We apply the proposed MCTS procedure to diverse benchmarks of math and commonsense reasoning tasks on both pretrained and instruction-finetuned LLMs. Surprisingly, the simple MCTS consistently finds higher-quality and/or shallower \ours for most samples, without training any parameters. This demonstrates the broad existence of better \ours architectures than the pretrained one for individual samples, underscoring the unexplored, compositional generalization of layers from pretrained models as separate modules. Moreover, we quantitatively evaluate their corrected errors and the reduced depths, highlighting a significant improvement in the depth-quality trade-off space. Furthermore, we conduct fine-grained analyses of the utilization of each layer in MCTS-optimized \ours and the impact of task difficulty on the depth of \ours, which shed novel insights into the redundancy and alignment of model architecture to specific tasks. Our key findings and main contributions can be summarized as:\looseness-1
\begin{itemize}[leftmargin=*]
    \item \textbf{We introduce a new dimension of generalization that turns a static pretrained LLM into dynamic architectures of adaptive depths without training any parameter}: for different test samples/tasks, the pretrained layers can be skipped, repeated, and assembled to create better (more accurate and/or shallower) \ours models without further training. 
    \item \textbf{We develop an MCTS protocol for efficient architecture search of \ours with adaptive depth} for each sample. In-depth analysis of patterns in the achieved \ours models sheds critical insights into the \textbf{importance and redundancy of layers} at different depths of pretrained/finetuned models of different sizes, which also vary for tasks at different difficulty levels. 
\end{itemize}

\section{Related Work}

\textbf{Layer Pruning and Early-Exit Neural Networks}
Many works aim to accelerate large Transformers by statically pruning weights or dynamically halting computation. Static pruning typically removes redundant neurons, heads, or layers after training. For example, \cite{liu2021ebert} demonstrate that a significant fraction of BERT's attention heads can be dropped with negligible performance loss, and \cite{gordon2020compressing} investigate fine-grained weight pruning in BERT. \cite{fan2019reducing} introduce LayerDrop, a structured dropout technique that effectively trains models so arbitrary subsets of layers can be skipped during inference without requiring fine-tuning. 
These methods produce smaller models that trade computation for a small accuracy loss.\looseness-1

By contrast, early-exit or input-adaptive methods add auxiliary classifiers at intermediate layers so that "easy" inputs exit early. Notable examples include FastBERT \cite{liu2020fastbert} and DeeBERT \cite{xin2020deebert}, which insert classifiers after each block and use confidence or entropy metrics to decide when to stop. PABEE \cite{zhou2020bert} employs a patience criterion to halt when predictions stabilize. DACT-BERT \cite{eyzaguirre2021dact} adopts a differentiable Adaptive Computation Time mechanism to learn how many Transformer layers to run for each example. \cite{liu2021faster} estimate input "hardness" via mutual information or reconstruction error to pre-determine the number of Transformer layers to use.

These early-exit networks achieve significant speedups on NLP tasks by adaptively reducing depth per input. More recently, early-exit ideas have been extended to vision and multimodal Transformers. \cite{xu2023lgvit} propose LGViT, which adds heterogeneous exit heads (local and global) to ViT so that vision transformers can terminate early with minimal feature loss. \cite{tang2023you} introduce MuE ("Multiple Exiting"), a strategy for unified vision-language models that dynamically skips layers in both encoder and decoder based on input similarity. These works demonstrate that later layers can be skipped to allow image and vision-language models to adapt computation per sample with minimal accuracy drop. Our work generalizes this approach by allowing skipping of arbitrary layers and enabling reuse of certain layers.\looseness-1

\textbf{Looped Transformer and Recurrent Depth}
Another line of research makes Transformer depth adaptive by looping or repeating layers. The Universal Transformer \cite{dehghani2018universal} was an early example: it applies the same self-attention block recurrently and uses a halting mechanism to determine when each position is "done" (adapting depth per token). Building on these ideas, recent work explicitly introduces loops in model architectures. \cite{fan2024looped} demonstrate that a Looped Transformer – a single Transformer block applied repeatedly – can achieve much better length generalization on algorithmic tasks by adjusting the number of loops during inference. Similarly, \cite{yang2023looped} note that looped architectures excel at learning algorithms by explicitly incorporating iterative characteristics into the transformer architecture. More sophisticated variants like the Inner Thinking Transformer \cite{chen2025inner} interleave adaptive loops with residual "thinking" connections and per-token routing, enabling the model to devote extra computation only to particularly difficult tokens. In summary, these approaches explore recurrent or elastic depth via explicit loops to tailor the number of applied layers to each input's complexity. Unlike our approach, they require special architecture design and training from scratch, whereas our work focuses on pure test-time adaptation.

\textbf{Dynamic Routing and Modular Inference}
A third theme treats networks as collections of modules or experts with dynamically chosen pathways per sample. Mixture-of-Experts (MoE) Transformer layers are a well-known example: they maintain multiple sub-networks ("experts") and route each token to a subset. \cite{wu2024routing} introduce Routing Experts (RoE) for multimodal LLMs, retrofitting trained models into a mixture-of-experts style by learning input-dependent shortcut routes through layers, guided by sparsity regularizers. \cite{jain2024mixture} present Mixture of Nested Experts (MoNE): experts organized in a hierarchy of increasing capacity, where tokens are sent to smaller experts when sufficient. MoNE learns to prioritize easy tokens through low-cost experts and reserve full models for hard cases, halving inference compute on ImageNet/Video tasks.

These methods exemplify sample-wise routing: at inference time, the model conditionally activates different sub-modules or experts for each input. Similarly, neural module networks \cite{andreas2016neural} assemble task-specific computation graphs from a library of modules. In modern LLMs/VLMs, these routing approaches – whether through gating experts, skipping layers, or assembling modules – form a spectrum of modular inference techniques that adapt the computation graph on a per-sample basis to balance cost and accuracy. Interestingly, our work suggests that transformer layers can function effectively as modules even without being specifically trained for that purpose.

\section{MCTS for Optimizing \ours with Adaptive Depths}

Our approach, \textbf{C}hain-\textbf{o}f-\textbf{La}yers (\ours), treats a pretrained LLM's layers as building blocks to be composed dynamically per input.
Formally, if the original model has layer $\left[L_1, \dots, L_N\right]$ in order, \ours seeks a new sequence (path) $P = \left[L_i, L_j, L_k, \dots\right]$ (each $L$ is one of the original layers) that can replace the standard forward pass for a given input.
The path can skip layers (omit some $L_m$) or loop layers (repeat some $L_m$ multiple times in succession, akin to unrolling an RNN).
We restrict that the path uses layers from the original model (no new weights) and each layer's internal parameters remain fixed – we only change the order and frequency of application.
In effect, \ours builds a custom ``sub-network'' out of the existing layers for each query.
This allows shallow execution for easy queries and deeper or iterative execution for hard queries, all within the same model's capacity.

\vspace{-8pt}
\begin{algorithm}[t]
\caption{Adaptive MCTS for \ours Optimization}
\label{alg:mcts}
\begin{algorithmic}[1]
\State Initialize root node $P_0 = [L_1, L_2, \dots, L_N]$
\For{$N = 1$ to number of simulations}
    \State \textbf{Selection}: traverse tree maximizing UCB($P$)
    \State \textbf{Expansion}: generate skip/repeat candidates if node unexplored
    \State \textbf{Simulation}: evaluate path accuracy on held-out input(s)
    \State \textbf{Backpropagation}: update $Q(P)$ and $v(P)$ along trajectory
\EndFor
\State \Return Pareto-optimal paths (accuracy vs. length)
\end{algorithmic}
\end{algorithm}

\vspace{0.5em}
\noindent\textbf{Search via Monte Carlo Tree Search.}
The space of possible layer paths is combinatorially large, especially when allowing both skips and repeats. To explore this space efficiently, we frame path construction as a symbolic search problem and use MCTS to discover optimal execution plans per input. Unlike greedy or beam search, MCTS better balances local exploitation with global exploration—crucial when optimal paths are long or non-trivial.
We formalize each MCTS game as follows:\looseness-1

\textbf{State.} A state is a partial or complete layer sequence, initialized as the standard forward path $P_0 = [L_1, L_2, \dots, L_N]$. At each step, the state evolves by applying a skip or repeat transformation.

\textbf{Actions.} Each action modifies a contiguous block of layers: skipping $k$ layers or repeating a block of $k$ layers $r$ times, where $k, r \in \{1,2,3,4\}$. For example, ``skip 2 layers'' removes the next 2, while ``repeat 3 layers twice'' inserts two additional copys of the next 3. We constrain the final path length to avoid excessive computation.

\textbf{Transition and Simulation.} A path is complete once no further actions are allowed. The path is executed on the input, yielding a prediction. A reward of 1 is assigned for a correct output, optionally minus a penalty proportional to the normalized path length to encourage compact solutions.

\textbf{Search Objective.} We define the UCB score for node selection as:
\[
\text{UCB}(P) = \underbrace{\frac{Q(P)}{v(P)}}_{\text{Exploitation}} + c \sqrt{\frac{\ln V}{v(P)}} - \underbrace{\lambda \frac{||P||}{N}}_{\text{Length Penalty}}
\]
where $Q(P)$ is the cumulative reward for path $P$, $v(P)$ is the visit count, $V$ is the total simulations, $c$ balances exploration, $\lambda$ scales the path-length penalty, and $||P||$ is the number of layers used.

We run a fixed number of simulations per input to explore the space. After search, we return the Pareto-optimal paths balancing accuracy and efficiency. The procedure is summarized in Algorithm~\ref{alg:mcts}.

\section{Experimental Setup}

We evaluate \ours across diverse pretrained and instruction-tuned LLMs, assessing its impact on generalization and computational efficiency. By enabling dynamic layer composition, \ours provides test-time flexibility beyond fixed-depth forward passes.

\textbf{Models.} We study three model families to examine how architecture and supervision influence dynamic depth selection:
\begin{itemize}[leftmargin=*]
    \item \textbf{LLaMA-3-3B, 8B:} Standard dense decoder-only transformers trained on open-domain corpora.
    \item \textbf{OLMoE-1B-7B:} A Mixture-of-Experts transformer with conditional routing. Each MoE layer is treated as a unit for skipping or repeating in \ours.
    \item \textbf{Instruction-tuned LLMs:} For all models above, we include their instruction-tuned versions to study the effect of supervised alignment on layer utility.
\end{itemize}

\textbf{Datasets.} We evaluate on two benchmark families that cover a broad spectrum of reasoning complexity: 1) ARC-E\textcolor{gray}{asy} and ARC-C\textcolor{gray}{hallenge}~\cite{clark2018think}: Commonsense reasoning tasks requiring shallow symbolic or factual inference; 
2) DART-Math~\cite{tong2024dartmath}: a math benchmark stratified into five difficulty levels (DART-1 to DART-5, from easiest to hardest), enabling analysis of how \ours adapts execution depth to task complexity.

\textbf{Evaluation Protocol.} All models are evaluated in a zero-shot setting with frozen weights. Each input is prompted as a natural language question, and model predictions are compared against gold answers. These outputs serve as feedback for MCTS to iteratively search optimized layer paths per input.

These experiments assess whether \ours can discover more efficient or more accurate computation paths than the default model, revealing the latent composability of pretrained LLM layers.

\section{Experimental Analysis}
\label{sec:exp_analysis}

To evaluate how compositional layer execution affects model behavior, we conduct a comprehensive empirical analysis across datasets, model scales, and training configurations. Our experiments aim to answer two core questions: (1) \textit{Does dynamic composition of pretrained layers improve generalization and efficiency?} and (2) \textit{How are individual layers engaged under adaptive execution paths?}

\subsection{Layer Composition in \ours Enhances Both Generalization and Efficiency}

A central hypothesis of this work is that fixed-depth computation can be a limiting factor for generalization and efficiency. By enabling test-time composition of layers—selectively skipping or repeating layers from a pretrained backbone—\ours offers a flexible alternative to the rigid forward pass of standard transformers. This section demonstrates that such dynamic layer composition substantially improves both prediction accuracy and computational efficiency across model scales and task types.\looseness-1

\begin{table}[t]
\vspace{-20pt}
\centering
\caption{\textbf{Comparing different architecture search spaces for layer composition.} Accuracy (\%) of the original LLM and the one achieved by MCTS in three different search spaces: skip-only, recurrence-only, or combining both (default of \ours). Our skip+recurrence joint search space consistently outperforms other two and the original LLM, especially on harder datasets such as DART-3/4/5.\looseness-1}
\label{tab:overall_accuracy}
\resizebox{\textwidth}{!}{
\begin{tabular}{llccccccc}
\toprule
\textbf{Model} & \textbf{Variant} & \textbf{ARC-E} & \textbf{ARC-C} & \textbf{DART-1} & \textbf{DART-2} & \textbf{DART-3} & \textbf{DART-4} & \textbf{DART-5} \\
\midrule
\multirow{4}{*}{\makecell[l]{LLaMA-3-3B-\\Base}}

  & Original         & 27.80 & 20.80 & 7.00 & 5.40 & 3.00 & 2.60 & 1.40  \\
  & +Skip-only         & \underline{75.40}  & \underline{75.60}  & \underline{27.80}  & \underline{18.40}  & \underline{18.40}  & \underline{13.00}  & \underline{8.20}  \\
  & +Recurrence-only        & 65.40  & 54.80  & 26.60  & 16.80  & 11.20  & 8.20  & 5.00  \\ %
  & +\ours (Skip+Recurrence)& \textbf{95.80} & \textbf{98.20} & \textbf{63.60} & \textbf{46.80} & \textbf{34.80} & \textbf{26.00} & \textbf{25.60} \\
\midrule
\multirow{4}{*}{\makecell[l]{LLaMA-3-3B-\\Instruct}}
  & Original        & 39.80 & 37.40 & 13.00 & 7.40 & 5.80 & 9.80 & 14.40 \\
  & +Skip-only             & \underline{86.80}  & \underline{86.60}  & 30.20  & 22.20  & 19.60  & 16.20  & 13.20  \\
  & +Recurrence-only           & 76.20  & 72.60  & \underline{40.60}  & \underline{25.20}  & \underline{21.00}  & \underline{18.20}  & \underline{23.80}  \\ %
  & +\ours (Skip+Recurrence)& \textbf{95.80} & \textbf{98.20} & \textbf{81.00} & \textbf{64.00} & \textbf{47.60} & \textbf{42.40} & \textbf{44.00} \\
\midrule
\multirow{4}{*}{\makecell[l]{LLaMA-3-8B-\\Base}}

  & Original        & 45.60 & 42.60 & 13.40 & 6.80 & 4.20 & 2.20 & 0.80 \\
  & +Skip-only        & \underline{89.20}  & \underline{86.60}  & \underline{39.80}  &  \underline{29.00} & \underline{20.00}  & \underline{16.40}  & \underline{12.20}  \\
  & +Recurrence-only       & 73.20  & 69.00  & 33.40  & 17.80  & 8.40  & 8.80  & 5.60  \\ %
  & +\ours (Skip+Recurrence) & \textbf{95.80} & \textbf{98.20} & \textbf{79.80} & \textbf{58.00} & \textbf{42.80} & \textbf{31.80} & \textbf{29.20} \\
\midrule
\multirow{4}{*}{\makecell[l]{LLaMA-3-8B-\\Instruct}}
  & Original        & 76.00 & 69.00 & 10.00 & 6.00 & 6.80 & 12.00 & 12.40 \\
  & +Skip-only        & \underline{92.80}  & \underline{93.20}  & \underline{44.20}  & \underline{25.20}  & \underline{25.00}  & 17.80  & 13.40  \\
  & +Recurrence-only       & 89.20  & 87.20  &  30.00 & 20.80   & 24.20  & \underline{24.40}  & \underline{21.60}  \\ %
  & +\ours (Skip+Recurrence)& \textbf{95.80} & \textbf{98.20} & \textbf{84.20} & \textbf{66.20} & \textbf{54.60} & \textbf{49.40} & \textbf{47.80} \\

\midrule
\multirow{4}{*}{\makecell[l]{OLMoE-1B-7B-\\Base}}
  & Original        & 24.80 & 21.40 & 12.60 & 3.40 & 3.00 & 1.80 & 1.80 \\
  & +Skip-only        & \underline{76.80} & \underline{77.60} & 26.00 & 15.20 & \underline{15.60} & \underline{8.80} & \underline{4.20} \\
  & +Recurrence-only       & 49.80 & 51.40 & \underline{30.20} & \underline{15.80} & 8.00 & 5.40 & 3.60 \\
  & +\ours (Skip+Recurrence)& \textbf{95.80} & \textbf{97.40}  & \textbf{57.60} & \textbf{41.20} & \textbf{32.60} & \textbf{23.00} & \textbf{16.80} \\
\midrule
\multirow{4}{*}{\makecell[l]{OLMoE-1B-7B-\\Instruct}}
  & Original        & 53.20 & 41.60 & 14.80 & 7.20 & 3.00 & 1.20 & 0.60 \\
  & +Skip-only        & \underline{89.40} & \underline{85.40} & 29.80 & 20.40 & \underline{18.60} & 9.00 & \underline{5.60} \\
  & +Recurrence-only       & 81.40 & 71.20 & \underline{36.80} & \underline{23.60} & 11.60 & \underline{9.60} & 5.00 \\
  & +\ours (Skip+Recurrence)& \textbf{95.80} & \textbf{98.00} & \textbf{63.80} & \textbf{48.00} & \textbf{36.80} & \textbf{28.00} & \textbf{22.00} \\
\bottomrule
\end{tabular}
}
\vspace{-12pt}
\end{table}

\begin{findingbox}[title={Finding 1}]
\textit{\textbf{Joint search of layer-skip and layer-recurrence significantly improves generalization.}} 
While layer-skip simplifies the architecture for easy inputs, and recurrence improves reasoning on moderate tasks, their combination consistently performs the best on the hardest tasks.\looseness-1
\end{findingbox}

To understand how different execution strategies impact generalization, we compare three search space variants: skip-only (allowing layers to be bypassed but used only once), recurrence-only (allowing repetition without skipping), and our full version (supporting both skipping and recurrence). 
This combination is motivated by the complementary effects of the two operations: \textit{skipping} allows the model to compress its computation, effectively pruning away redundant or less informative layers, while \textit{recurrence} enables expansion of depth by revisiting high-utility layers for iterative refinement. When used together, they grant the model both compression and expansion capabilities—allowing it to adaptively match the depth and structure of reasoning to the input. This flexibility forms the foundation of \ours's generalization gains across diverse tasks.

As shown in Table~\ref{tab:overall_accuracy}, both restricted variants improve performance over vanilla forward execution, but in complementary and ultimately limited ways.

The \textit{skip-only variant} provides strong gains on simpler tasks, particularly ARC-E and ARC-C, suggesting that many inputs can be correctly processed with shallower depth. For example, on LLaMA-3B, accuracy on ARC-E improves from 27.8\% to 75.4\%, showing that reducing computation depth can be beneficial when the full model is over-parameterized for the task.

The \textit{recurrence-only variant} tends to perform better on more complex tasks that benefit from repeated application of informative layers. For instance, on DART-4 and DART-5, recurrence-only outperforms significantly skip-only in several settings, including LLaMA-3-3B-Instr and LLaMA-3-8B-Instr, where simply shortening the path is insufficient to capture task-specific dependencies.

However, \textbf{neither skip-only nor recurrence-only suffices across the board}. Our full method, which searches over both skipping and recurrence decisions, consistently achieves the highest accuracy on all datasets and models. These gains are especially large on harder DART-Math datasets: for example, on DART-2, the LLaMA-3-8B-Instr model improves from 25.2\% (skip-only) and 22.0\% (recurrence-only) to 66.2\% with the full space—nearly a threefold increase.

Moreover, we observe a consistent pattern across the MoE model OLMoE as well. Although the absolute improvement from \ours is somewhat smaller compared to dense models, this is likely due to the fact that OLMoE already employs sparse expert selection at each layer, leaving less room for further compression or expansion. Nevertheless, \ours is still able to discover substantially better execution paths on OLMoE, highlighting that even sparse architectures benefit from dynamic layer composition when equipped with flexible skipping and recurrence operations.

\begin{findingbox}[title={Finding 2}]
\textit{\textbf{Reducing the depth helps correct errors.}}
Various input tasks can be solved correctly using shallower and highly compressed layer compositions. \ours's error correction often leads to even fewer layers than keeping originally correct predictions, indicating severe ``overthinking'' in depth. \looseness-1
\end{findingbox}

\begin{figure}[h]
    \begin{center}
    \vspace{-12pt}
    \includegraphics[width=1.0\textwidth]{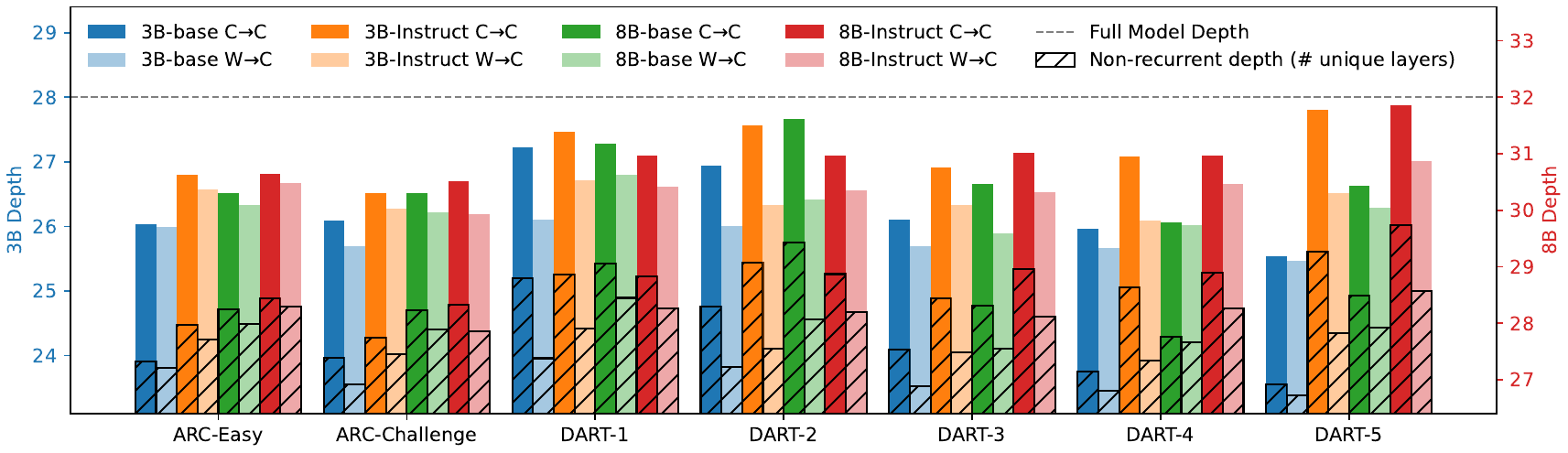}
  \end{center}
  \caption{\textbf{Depth and non-recurrent depth (\# unique layers) of \ours on four models and seven benchmarks.} The average depth of \ours for inputs whose predictions by the original model and \ours are both correct (\textsc{C$\rightarrow$C}), and whose predictions by the original model are wrong but corrected by \ours (\textsc{W$\rightarrow$C}). Both the depth and non-recurrent depth are effectively reduced in all cases. 
}
\vspace{-8pt}
  \label{fig:avg_length}
\end{figure}

Beyond accuracy, we investigate whether these gains are achieved with lower inference cost.
Figure~\ref{fig:avg_length} analyzes the average  depth under \ours across all tasks and models, revealing consistent and substantial gains in inference efficiency.

We observe that, for both \textsc{C$\rightarrow$C} (correct before and after) and \textsc{W$\rightarrow$C} (corrected by \ours) inputs, the average depth is significantly lower than the full model depth. This confirms that many inputs can be correctly processed without invoking the entire network—underscoring the potential for computation-aware inference.

Moreover, the distinction between total depth and non-recurrent depth (i.e., the number of unique layers used) reveals the extent of \textbf{layer recurrence}. In many cases, non-recurrent depth is markedly lower than total depth, indicating that \ours reactivates high-utility layers multiple times to achieve correct answers more compactly. This recurrence-driven compression is most pronounced on simpler datasets such as ARC-E, but remains visible even on harder DART tasks.

Interestingly, \textsc{W$\rightarrow$C} paths tend to be shorter than \textsc{C$\rightarrow$C} ones—both in total and non-recurrent depth. This suggests that correcting an incorrect prediction does not require more computation, but rather a reconfiguration that omits noisy or misleading layers present in the default forward pass.

Finally, we find that pretrained models generally yield more compressed execution paths than their instruction-tuned counterparts. This is likely because instruction tuning calibrates more layers to be relevant, leaving less opportunity for aggressive pruning or recurrence. In contrast, pretrained models often contain layers that can be skipped or consolidated, allowing \ours to discover leaner and more effective execution patterns.

\begin{wrapfigure}{l}{0.48\textwidth}
  \vspace{-2.em}
  \centering
  \includegraphics[width=0.46\textwidth]{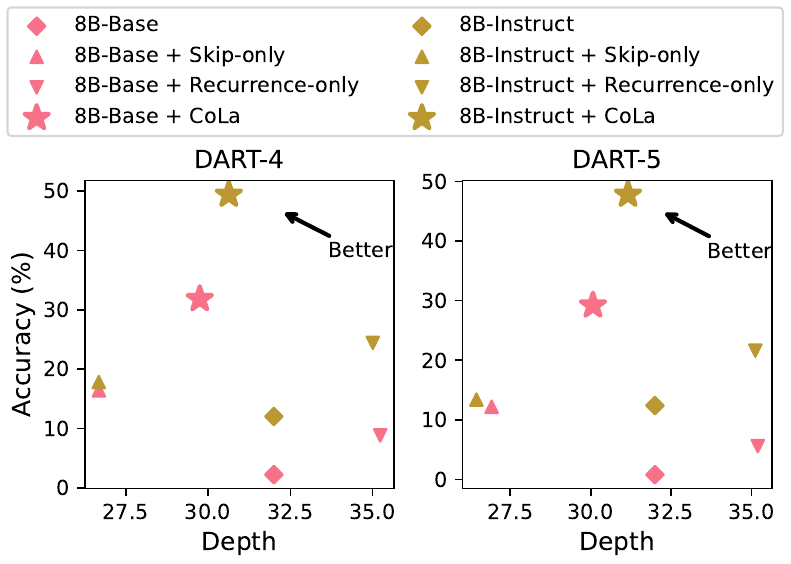}
  \caption{\textbf{Accuracy–depth tradeoff} on DART-4/5 (hardest). Each point represents an architecture search space (original, skip-only, recurrence-only, \ours) applied to pretrained or instruction-tuned LLaMA-3-8B. \ours consistently achieves the best tradeoff: although its depth lies between skip-only and recurrence-only strategies, it substantially improves accuracy, pushing the accuracy–cost Pareto frontier forward.\looseness-1} %
\label{fig:accuracy_depth_tradeoff}
  \vspace{-2.em}
\end{wrapfigure}

These patterns are further illustrated in Figure~\ref{fig:accuracy_depth_tradeoff}, which compares the actual depth and accuracy of each strategy on DART-4 and DART-5. Recurrence-only strategies achieve moderate gains over the pretrained baseline, but do so by increasing the overall depth—relying on repeated application of helpful layers to compensate for rigid forward computation. In contrast, skip-only strategies operate with reduced depth by omitting irrelevant layers, yielding marginal improvements in efficiency and accuracy. Strikingly, \ours achieves substantially higher accuracy than both skip-only and recurrence-only strategies, despite not always using the fewest layer invocations. Its ability to balance depth and performance highlights the strength of test-time dynamic layer composition—achieving superior generalization with efficient yet targeted computation.\looseness-1

\begin{findingbox}[title={Finding 3}]
\textit{\textbf{Keeping predictions correct does not require the full-depth model and all the layers.}} Even for inputs whose predictions by the original model are already correct, \ours can find shallower and more effective architectures, revealing substantial redundancy in static transformer inference.
\end{findingbox}

\begin{figure}[h]
    \begin{center}
    \vspace{-18pt}
    \includegraphics[width=1.0\textwidth]{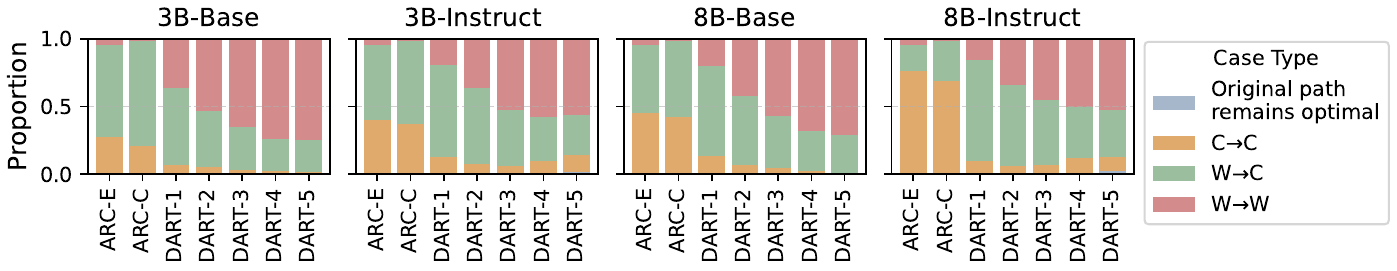}
  \end{center}  
  \vspace{-6pt}
  \caption{\textbf{Prediction correctness transitions under \ours.} For each model and dataset, it reports the proportion of four categories of test samples: original path remains optimal, correct$\rightarrow$correct (\textsc{C$\rightarrow$C}), wrong$\rightarrow$correct (\textsc{W$\rightarrow$C}), and wrong$\rightarrow$wrong (\textsc{W$\rightarrow$W}). \ours substantially improves prediction outcomes. It rarely retains the original sub-optimal path. }
  \vspace{-15pt}
  \label{fig:case_transition1}
\end{figure}

We further analyze how \ours reshapes prediction outcomes at the example level. Figure~\ref{fig:case_transition1} categorizes test inputs into four transition types: \textsc{C$\rightarrow$C}, \textsc{W$\rightarrow$C}, wrong before and after (\textsc{W$\rightarrow$W}), and cases where the original path remains optimal.

Across all models and datasets, the fraction of inputs where the original path remains optimal is nearly negligible—even among those correctly solved without \ours. This reveals that static forward passes are rarely optimal, and that \ours consistently discovers alternative paths that lead to the same or better outcomes. A substantial portion of inputs fall into the \textsc{W$\rightarrow$C} category, demonstrating \ours's strong capacity to recover from errors. While this corrective ability diminishes slightly on the hardest benchmarks (e.g., DART-5), gains remain significant across settings.\looseness-1

These results show that \ours not only improves accuracy through path reconfiguration, but does so more efficiently—often correcting errors or simplifying computation on a per-input basis. This challenges the assumption that correct predictions imply optimal computation, and underscores the benefit of flexible, structure-aware inference.\looseness-1

\subsection{Layer Engagement: Selection, Skipping, and Recurrence}
To better understand how \ours executes depth-adaptive computation, we study how different layers are engaged across tasks and model scales. We organize the analysis into two parts: (1) which layers are selected, and (2) how layers are recurred or skipped.

\begin{findingbox}[title={Finding 4}]
\textit{\textbf{Harder tasks and larger models encourage more uniform layer engagement.}} As tasks become more challenging and model capacity increases, \ours distributes computation more evenly across layers—moving from shallower specialization to deeper models involving more diverse layers.\looseness-1
\end{findingbox}

\begin{figure}[h]
    \vspace{-12pt}
    \begin{center}
    \includegraphics[width=1.0\textwidth]{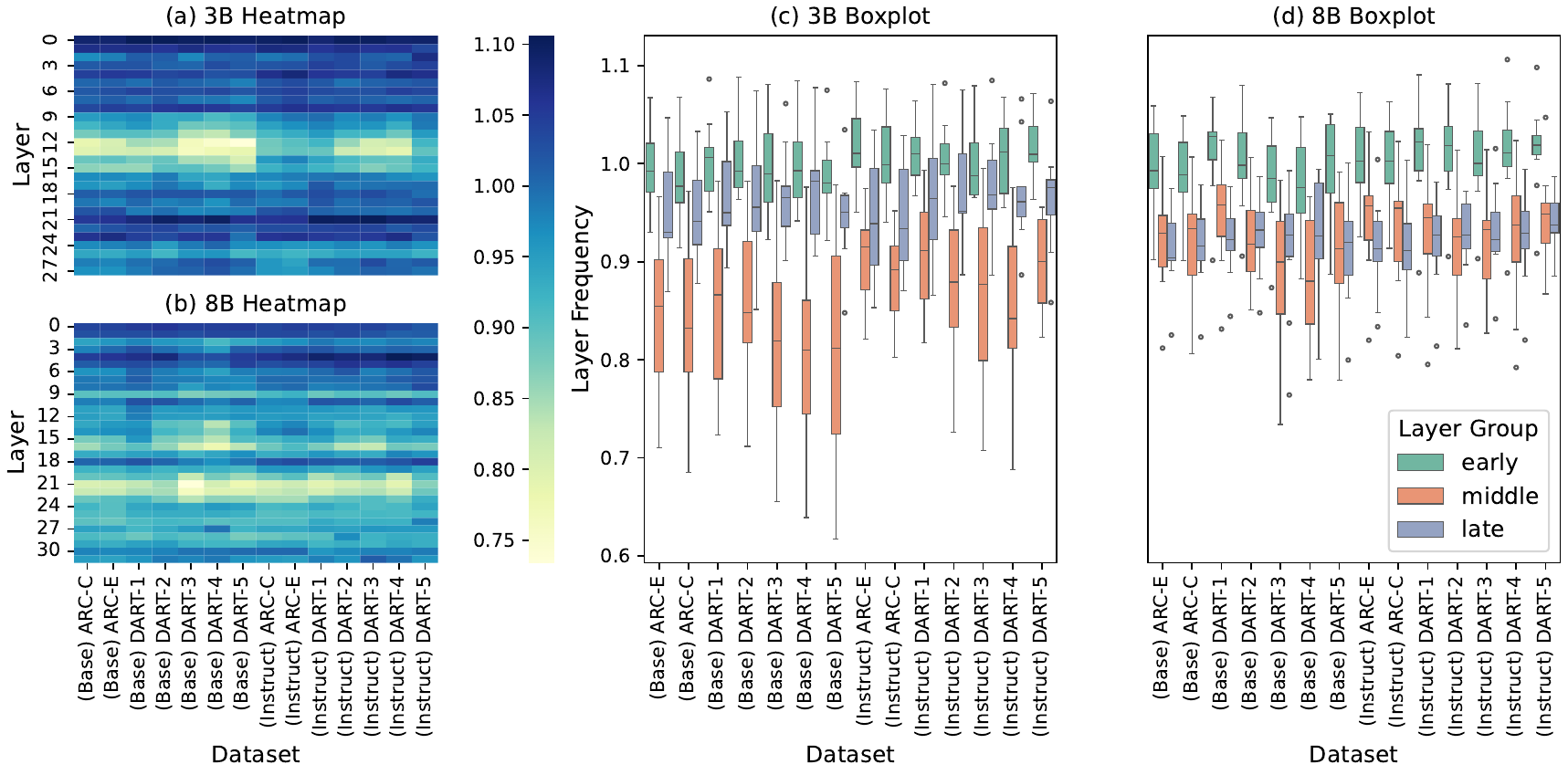}
  \end{center}
  \vspace{-4pt}
  \caption{\textbf{Layer selection patterns in \ours.}
  (a, b) Heatmaps show the frequency of each layer being selected for 3B and 8B models, respectively, on each dataset, with darker shades indicating higher usage.
  (c, d) Boxplots group layers into early, middle, and late segments, revealing systematic variation in their usage across datasets and models. 3B models exhibit greater variability and more aggressive pruning of mid-depth layers compared to 8B.
  \looseness-1}
  \vspace{-12pt}
  \label{fig:layer_freq}
\end{figure}

\textbf{Layer selection patterns reflect task complexity and depends on model scales.} 
We analyze how layer usage varies with model scale and task difficulty (Figure~\ref{fig:layer_freq}). Across all settings, early layers are most frequently engaged, likely due to their broad utility in extracting low-level features. However, deeper patterns diverge between models.\looseness-1

In the 3B model, layer usage follows a clear V-shape: early and late layers dominate, while middle layers are suppressed. This suggests a tendency to compress computation into the extremes of the network. In contrast, the 8B model exhibits a smoother decay from early to late layers, with middle layers more consistently engaged—reflecting greater capacity to leverage intermediate representations. Larger models also exhibit more stable usage patterns, with reduced variance and fewer outliers, especially after instruction tuning.
Quantitatively, 8B models show greater diversity in layer engagement, with higher entropy (3.46 v.s. 3.33) and lower maximum layer concentration (0.035 v.s. 0.040) compared to 3B. This indicates less reliance on any single layer and a more distributed use of the model’s depth. %

Task complexity further amplifies these differences. On harder datasets such as DART-5, layer usage becomes increasingly uniform, especially in 8B models. Early, middle, and late layers are engaged with comparable frequency, suggesting that complex reasoning benefits from activating a broader slice of the model’s capacity.

\begin{findingbox}[title={Finding 5}]
\textit{\textbf{Larger finetuned models adopt less stereotyped layer-usage patterns.}} While smaller models' \ours show skipping and recurrence patterns concentrated on layers at specific depths, larger and instruction-tuned models engage layers more irregularly and adaptively regardless of their depths.
\end{findingbox}

\begin{figure}[h]
\vspace{-4pt}
    \begin{center}
    \includegraphics[width=1.0\textwidth]{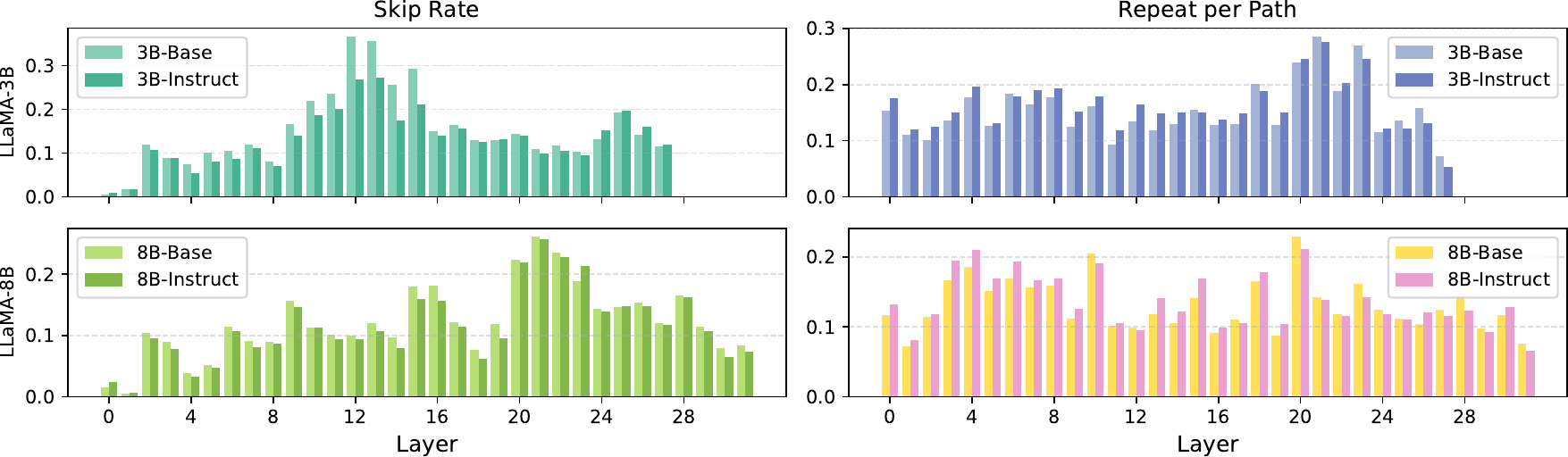}
  \end{center}
  \vspace{-4pt}
  \caption{\textbf{Skipping and recurrence rate of each layer on four models.}
  Left: Skip rate—the proportion of \ours in which layer-$i$ is skipped. Right: Averaged recurrence times of layer-$i$ in \ours. \ours models consistently keep early layers but exhibit an elevated skip rate of middle layers.\looseness-1
  }
  \vspace{-4pt}
  \label{fig:case_transition2}
\end{figure}

\textbf{Fine-Grained Analysis of Skipping and Recurrence Patterns.}
To gain finer-grained insights into how layers are used during inference, we examine not only whether a layer is selected, but how often it is skipped entirely or recurred multiple times within a path. Figure~\ref{fig:case_transition2} visualizes the per-layer \textit{skip rate} and \textit{repeat count per path}, aggregated over all datasets and grouped by model variant.\looseness-1

Across all models, the earliest layers are consistently retained during inference, exhibiting near-zero skip rates. Beyond this point, skip rates follow a non-monotonic profile: increasing across intermediate layers before declining toward the deeper end. This pattern is most pronounced in smaller pretrained models, where skipping is strongly concentrated in a narrow middle region. Instruction tuning softens this behavior, resulting in a smoother and more dispersed skip distribution.
Larger models exhibit broader and less localized skipping patterns, with elevated skip rates spread across both middle and late layers. While they lack the sharp spikes seen in smaller models, their skip profiles still exhibit moderate fluctuations—indicating neither uniformity nor strict localization.

Recurrence distributions also show depth-dependent variation, though their profiles differ from skip patterns in where repetition is concentrated.
Smaller pretrained models concentrate repetition toward late layers, while tuning promotes more even recurrence across the stack. In larger models, recurrence counts fluctuate non-monotonically across depth, suggesting that layer recurrence is dynamically adjusted based on context rather than fixed to particular depths.

Together, these results suggest that smaller models rely on stereotyped usage—amplifying or pruning specific depths—while larger and instruction-tuned models adopt more distributed, context-sensitive computation paths.

\section{Conclusion}

This work reframes model generalization through test-time architectural adaptation. By treating each transformer layer as a reusable or skippable module, we introduce a flexible Chain-of-Layers (\ours) space that enables input-specific execution paths without additional training. Using Monte Carlo Tree Search, \ours discovers optimized layer compositions that improve both accuracy and efficiency across diverse tasks and models. Empirically, it not only corrects original model errors but often does so with shallower, more targeted computation. These results challenge the static nature of standard forward passes and reveal the latent compositionality of pretrained layers—positioning test-time depth adaptation as a promising step toward unifying fast and slow reasoning in LLM inference.

\bibliographystyle{plainnat}
\bibliography{references}

\clearpage

\appendix

\section{Fine-Grained Depth Compression Analysis}

\begin{figure}[h]
\centering
\includegraphics[width=\textwidth]{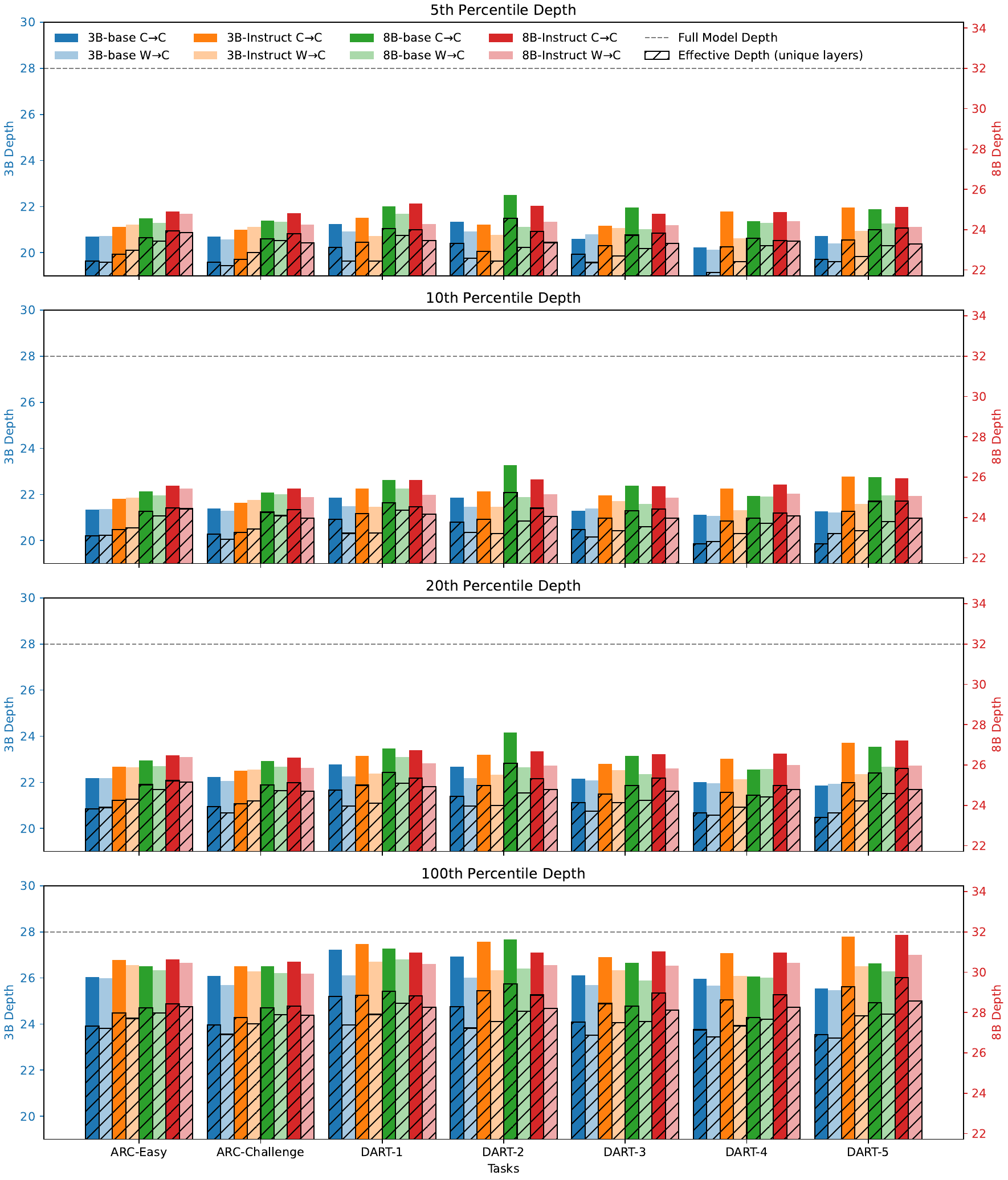}
\caption{\textbf{Mean depth and non-recurrent depth of the shortest 5\%, 10\%, 20\%, and 100\% of valid execution paths under \ours.}}
\label{fig:percentile_depth}
\end{figure}

To deepen our understanding beyond average-case trends, we conduct a percentile-based analysis of inference path lengths under \ours. Instead of aggregating over all inputs, we focus on the most efficient cases by identifying the shortest valid execution paths and computing their mean total and non-recurrent depth. For each model and dataset, we report the average depth among the shortest 5\%, 10\%, 20\%, and 100\% of \ours paths for correctly solved inputs, including both \textsc{C$\rightarrow$C} and \textsc{W$\rightarrow$C} transitions. The results, shown in Figure~\ref{fig:percentile_depth}, highlight how much computation can be reduced in the most efficient cases.

We analyze the mean total and non-recurrent depth of the shortest 5\% and 20\% of correct execution paths under \ours (Figure~\ref{fig:percentile_depth}). For the top 5\% most efficient cases, 3B models compress to 20--22 layers and 8B to 22.5--25, corresponding to \textbf{up to 30\% reduction} from the full model depth. For the 20th percentile, depths remain 12--23\% lower than baseline, indicating that a substantial fraction of inputs---especially corrected errors---can be processed with significantly fewer layers. Non-recurrent depths are consistently even lower, confirming the presence of efficient recurrence patterns and layer reuse in shallow yet effective execution paths.

\section{Implementation Details}

\textbf{Algorithm.}
We use \texttt{200} simulations per input in our MCTS implementation for optimizing \ours, balancing search quality and runtime. The UCB score incorporates a normalized path length penalty with weight \texttt{5.0} to favor compact execution paths. To encourage exploration, the algorithm selects a random unexplored child node with probability \texttt{0.1} instead of the one with the highest UCB score. This behavior is hard-coded as a fixed conditional in the selection logic.
All hyperparameters are kept fixed across models and datasets to ensure consistency and reproducibility.

\textbf{Datasets.}
We randomly sample \texttt{500} instances from each dataset for evaluation.

\end{document}